\documentclass[10pt,journal,compsoc]{IEEEtran}

\ifCLASSOPTIONcompsoc
  \usepackage[nocompress]{cite}
\else
  \usepackage{cite}
\fi
\ifCLASSINFOpdf
\else
\fi
\usepackage{amsmath}
\usepackage{amssymb}
\usepackage{bm}
\usepackage{graphicx}
\usepackage{multirow}

\makeatletter
\newcommand{\thickhline}{%
    \noalign {\ifnum 0=`}\fi \hrule height 1pt
    \futurelet \reserved@a \@xhline
}
\makeatother

\ifCLASSOPTIONcompsoc
  \usepackage[caption=false,font=footnotesize,labelfont=sf,textfont=sf]{subfig}
\else
  \usepackage[caption=false,font=footnotesize]{subfig}
\fi
\let\MYorigsubfloat\subfloat
\renewcommand{\subfloat}[2][\relax]{\MYorigsubfloat[]{#2}}
\usepackage{url}

\begin{document}
%
\title{SimVODIS: Simultaneous Visual Odometry, Object Detection, and Instance Segmentation}
%
%
%
%

\author{Ue-Hwan~Kim, Se-Ho~Kim
        and~Jong-Hwan~Kim,~\IEEEmembership{Fellow,~IEEE}
\IEEEcompsocitemizethanks{\IEEEcompsocthanksitem The authors are with the School of Electrical Engineering, KAIST (Korea Advanced Institute of Science and Technology), Daejeon, 34141, Republic of Korea.\protect\\
E-mail: \{uhkim, shkim, johkim\}@rit.kaist.ac.kr}
}

%
%

\markboth{IEEE Transactions on Pattern Analysis and Machine Intelligence}
{Shell \MakeLowercase{\textit{et al.}}: Bare Demo of IEEEtran.cls for Computer Society Journals}
%



\IEEEtitleabstractindextext{%
\begin{abstract}
Intelligent agents need to understand the surrounding environment to provide meaningful services to or interact intelligently with humans. The agents should perceive geometric features as well as semantic entities inherent in the environment. Contemporary methods in general provide one type of information regarding the environment at a time, making it difficult to conduct high-level tasks. Moreover, running two types of methods and associating two resultant information requires a lot of computation and complicates the software architecture. To overcome these limitations, we propose a neural architecture that simultaneously performs both geometric and semantic tasks in a single thread: simultaneous visual odometry, object detection, and instance segmentation (SimVODIS). Training SimVODIS requires unlabeled video sequences and the photometric consistency between input image frames generates self-supervision signals. The performance of SimVODIS outperforms or matches the state-of-the-art performance in pose estimation, depth map prediction, object detection, and instance segmentation tasks while completing all the tasks in a single thread. We expect SimVODIS would enhance the autonomy of intelligent agents and let the agents provide effective services to humans.
\end{abstract}

}

\maketitle

\IEEEdisplaynontitleabstractindextext

%
\IEEEpeerreviewmaketitle

\IEEEraisesectionheading{\section{Introduction}\label{sec:introduction}}

\IEEEPARstart{A}{s} technology advances, AI replaces simple repetitive and dangerous tasks, leading people to live a better and more convenient life. Understanding the surrounding environment is essential for an intelligent agent to provide meaningful services to or to interact with people \cite{kim20193}. In the process, the intelligent agent must be able to understand the semantic entities inherent in the environment as well as the geometry of the environment in order to perform high-level tasks such as errands, cleaning, cooking, and answering questions \cite{mccormac2018fusion++}. If the intelligent agent collects only one type of information among the physical information and semantics of the surrounding environment, it can only perform simple tasks and cannot provide meaningful services to humans.

Methods for the intelligent agent to understand the surrounding environment include extracting geometric information such as visual odometry (VO) \cite{forster2016svo, engel2018direct} and SLAM \cite{zou2013coslam, mur2017orb} and recognizing semantics such as object detection \cite{redmon2016you, he2017mask} and semantic mapping \cite{ma2017multi, mccormac2017semanticfusion}. Algorithms that collect only geometric information are merely suitable for simple tasks such as navigation and path planning because they lack semantic information to provide high-level services. On the other hand, methods for obtaining semantic information in general process one image at a time making it difficult to provide a practical service in the real-world contexts. Recently proposed approaches combine two techniques to associate geometric and semantic information \cite{bowman2017probabilistic, xiang2017darnn}. However, running both algorithms simultaneously requires a lot of computation resources and complicates the software structure \cite{zhong2018detect}.

\begin{figure}
    \centering
    \includegraphics[width=0.45\textwidth]{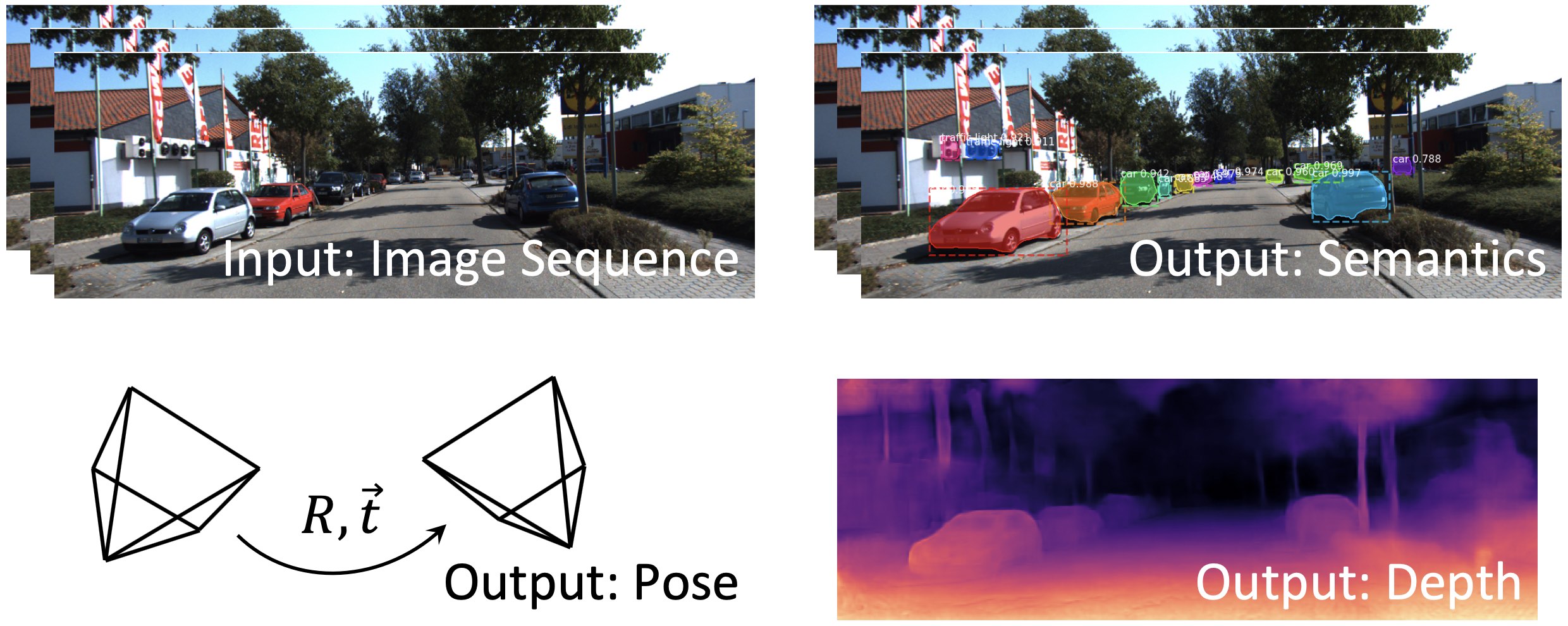}
    \caption{Overview of the proposed SimVODIS. SimVODIS receives a set of three consecutive images and then estimates semantics (object classes, bounding boxes and object masks), relative poses between input images, and the depth map of the center image.}
    \label{fig:overview}
\end{figure}

To overcome the above-mentioned limitations, we propose Simultaneous Visual Odometry, Object Detection, and Instance Segmentation (SimVODIS). SimVODIS is a neural network that concurrently estimates geometry and semantics of the surrounding environment (Fig. \ref{fig:overview}). SimVODIS evaluates the following information when it receives a set of image frames: 1) relative pose between image frames, 2) depth map prediction, 3) object classes, 4) object bounding boxes, and 5) object masks. To the best of our knowledge, SimVODIS is the first fully data-driven semantic VO algorithm. We expect SimVODIS would prompt the development of data-driven semantic VO/SLAM since data-driven VO algorithms would advance from geometric VO to semantic VO as conventional feature-based VO/SLAM have evolved.

We design the SimVODIS network on top of the Mask-RCNN architecture \cite{he2017mask}. Since Mask-RCNN derives common features to perform both geometric and semantic tasks, these features allow the design of a multi-task network that performs both geometric and semantic tasks. We devise two network branches utilizing the features for pose estimation and depth map prediction. Because SimVODIS is an unsupervised learning framework, training SimVODIS requires just unlabeled image sequences. In the training phase in which pretrained Mask-RCNN is employed, each input image frame is warped using the estimated relative pose and the predicted depth map, and training losses are calculated using the correspondence between the warped image and the original image.

Previous studies have mainly used the KITTI dataset \cite{geiger2013vision} which captures outdoor scenes to train the networks for joint estimation of ego-motion and depth map prediction. However, intelligent agents would work in both outdoor and indoor environments such as autonomous cars that run on the outside roads and park inside a building. We supplement a set of datasets and investigate the effect of the dataset heterogeneity on model performance in mixed outdoor and indoor scenarios. In addition, we analyze model performance according to various training conditions through the ablation study.

Specifically, the main contributions of our work are as follows.
\begin{enumerate}
\item \textbf{Research Scenario}: We define a fully data-driven semantic VO algorithm, SimVODIS, for the first time. We expect SimVODIS would provoke the evolution of data-driven VO towards semantic VO/SLAM.
\item \textbf{Network Architecture Design}: We design the SimVODIS network which simultaneously performs both geometric and semantic tasks. The network conducts multiple tasks utilizing shared feature maps and runs in one thread.
\item \textbf{Studying the Effect of Dataset Heterogeneity}: We employ multiple datasets for training the proposed SimVODIS and investigate the effect of dataset heterogeneity on the performance of ego-motion estimation and depth map prediction.
\item \textbf{Ablation Study}: We vary the training conditions in multiple ways and evaluate how different training environments affect the performance of the SimVODIS network.
\item \textbf{Open Source}: We contribute to the corresponding research society by making the source code of the proposed SimVODIS network and the pretrained network parameters public\footnote{\url{https://github.com/Uehwan/SimVODIS}}.
\end{enumerate}

The rest of this manuscript is structured as follows. Section II reviews previous research outcomes related to SimVODIS. Section III describes the proposed SimVODIS network and the training scheme. Section IV delineates the evaluation setting and Section V illustrates the evaluation results with corresponding analysis. Section VI discusses future research directions for further improvement of SimVODIS and concluding remarks follow in Section VII.
\section{Related Works}
\label{sec:related_works}
We review previous research outcomes relevant to the proposed SimVODIS in this section. We discuss the main ideas of previous works, compare them with SimVODIS and point out the novelty of SimVODIS.

\begin{figure*}[!t]
\centering
\subfloat[]{\includegraphics[width=0.9\textwidth]{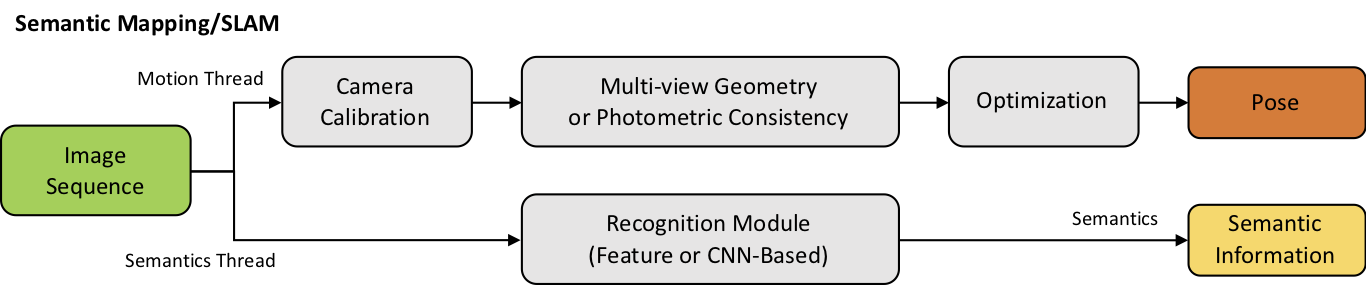}%
\label{fig:comparison_semantic_vo}}
\hfil
\subfloat[]{\includegraphics[width=0.9\textwidth]{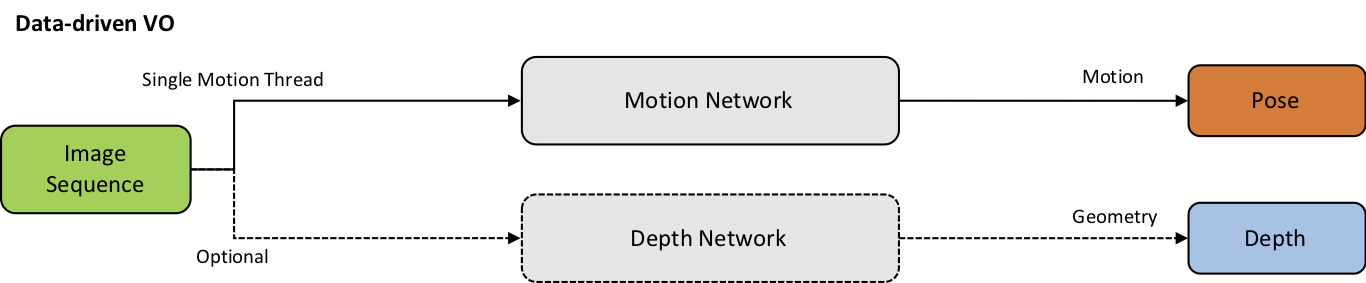}%
\label{fig:comparison_data-driven_vo}}
\hfil
\subfloat[]{\includegraphics[width=0.9\textwidth]{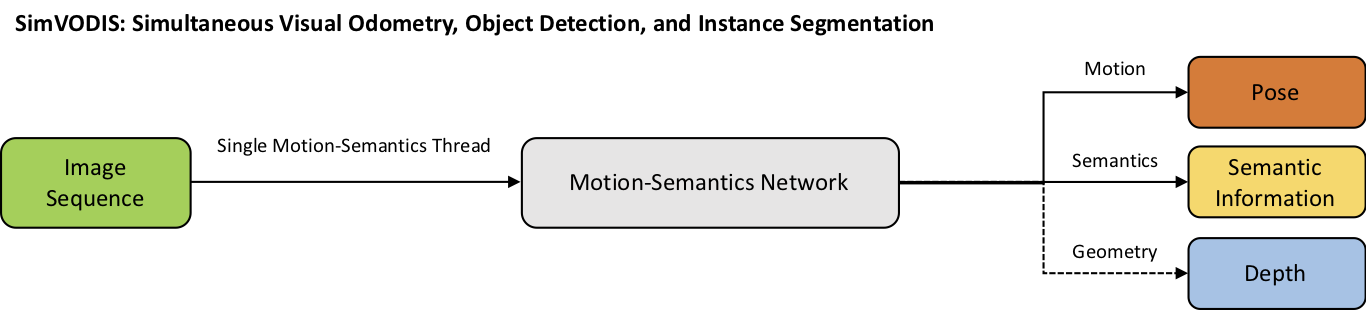}%
\label{fig:comparison_simvodis}}
\caption{SimVODIS compared to conventional methods. (a) Semantic Mapping/SLAM, (b) Data-driven VO, (c) Proposed SimVODIS}
\label{fig:comparison}
\end{figure*}

\subsection{Semantic Mapping/SLAM}
Conventional VO/SLAM methods effectively provide intelligent agents with physical information for pose estimation and navigation. However, intelligent agents equipped with conventional VO/SLAM methods lack any semantic information and require additional recognition processes for reasoning-enabled high-level AI applications. To enable intelligent agents to understand the semantics of surrounding environments and perform various tasks, researchers have attempted to associate semantics from recognition modules with physical information from VO/SLAM modules. For this, a semantics thread with a recognition module is added to the original motion thread to implement semantic Mapping/SLAM. The recognition module is either feature-based or CNN-based. Fig. \ref{fig:comparison_semantic_vo} shows the architecture of semantic Mapping/SLAM.

\subsubsection{Feature-based Methods}
Feature-based methods utilize hand-crafted features to recognize semantic entities. Most commonly, monocular cameras are employed since monocular cameras provide an inexpensive and easy-to-deploy way to collect visual information of the surrounding environment. Dense feature maps such as SIFT descriptors \cite{pillai2015monocular} are extracted from monocular video sequences or bags of binary words with a database of 3D object models \cite{galvez2016real} are exploited for recognition of objects or other entities. At the expense of cost and an additional calibration step, the usage of RGB-D cameras enhances the performance of semantic Mapping/SLAM compared to the methods with monocular cameras \cite{stuckler2015dense, scona2018staticfusion}, since the additional depth information makes the extracted features richer.

However, feature-based recognition modules are gradually giving their ways to CNN-based methods for a couple of reasons. First, feature-based methods display lower accuracy than deep learning based state-of-the-art methods \cite{lecun2015deep}. With the improved computation power and the availability of bigger datasets, the features learned by deep neural networks surpass the performance of hand-crafted features. Next, feature-based methods require manual work for designing features. Designing features takes much time and effort in addition to a lot of experience.

\subsubsection{CNN-based Methods}
CNN-based recognition modules are prevailing due to their high performance and CNN-based methods are replacing the feature-based recognition modules of semantic Mapping/SLAM methods. Current semantic Mapping/SLAM methods employ both off-the-shelf object classifiers \cite{sunderhauf2017meaningful, yu2018ds, mccormac2018fusion++} and fine-tuned modules \cite{mccormac2017semanticfusion, xiang2017darnn} to extract semantic information. Nonetheless, CNN-based methods suffer from increased system complexity and computation time, since such methods run in two threads: motion and semantic threads. The increased system complexity could lead to maintenance problem and the prolonged computation time could hinder real-time applications. The proposed SimVODIS, in contrast, run in one thread to extract both pose and semantic information, minimizing the growth of the system complexity and computation time.

\subsection{Data-Driven VO}
The recognition modules of semantic Mapping/SLAM are replaced with data-driven (CNN-based) methods as aforementioned. However, the VO (pose estimation) part of semantic Mapping/SLAM has not been completely replaced with data-driven methods, since the research on the data-driven VO is in the early stage at the moment. Researchers have started to show its feasibility \cite{zhou2017unsupervised} and the field is under active research. Data-driven VO can be categorized into two classes: supervised and unsupervised VO.

\subsubsection{Supervised VO}
The research on the data-driven VO has started with the supervised learning scheme. At first, researchers have formulated the pose estimation with deep neural networks (DNN) as a regression problem and trained DNNs to regress the pose of the current image frame for a re-localization purpose \cite{kendall2015posenet}. Later, DNNs have been trained to evaluate the relative pose between two consecutive image frames and showed satisfactory performance as an initial study \cite{ummenhofer2017demon, wang2018end}. Nonetheless, the research on the data-driven VO is moving towards the unsupervised VO due to a few limitations: Supervised VO requires labeled data for training and securing labeled data for VO necessitates the usage of extra sensor tools \cite{geiger2013vision}.

\subsubsection{Unsupervised VO}
Unsupervised VO does not require labeled data for training. It utilizes image reconstruction from different camera views as a supervision signal. Unsupervised VO jointly trains two CNNs, pose estimation and depth map prediction networks, since the image reconstruction needs both relative pose and depth map. Unsupervised VO deals with two types of input image sequences: monocular and stereo image sequences. The unsupervised VO with monocular image sequences \cite{zhou2017unsupervised, mahjourian2018unsupervised, godard2019digging} entails the scale ambiguity problem, while the unsupervised VO for stereo image sequences \cite{godard2017unsupervised, zhan2018unsupervised} requires a more expensive device and additional calibration steps. The proposed SimVODIS extends unsupervised VO with monocular image sequences because the development process for a SLAM system starts from a monocular VO and develops towards a stereo and rgb-d VO and a final SLAM system.

\subsubsection{Comparison with SimVODIS}
The contemporary data-driven VO methods, as conventional feature-based VO and SLAM in the early stage, only extract physical information and do not offer semantic information. Thus, intelligent agents cannot solely rely on the contemporary data-driven VO for versatile performance. In contrast, the proposed SimVODIS provides both physical and semantic information in one thread allowing intelligent agents to understand the surrounding environment in a deeper manner.
\section{SimVODIS}
In this section, we illustrate the proposed network architecture of SimVODIS built on top of Mask-RCNN and the framework for unsupervised learning of depth and ego-motion from monocular videos.

\subsection{Problem Formulation}
We represent ego-motion in $\mathbb{R}^3$ as a motion vector as follows:
\begin{equation}
    \bm{u}_{target, source} = [\bm{t}^T, \bm{r}^T]^T,
\end{equation}
where $\bm{t} = [\Delta x, \Delta y, \Delta z]^T$ is a translation vector and $\bm{r} = [\Delta \theta, \Delta \phi, \Delta \psi]^T$ is a rotation vector expressed in Euler angle. We express the rotation vector as an Euler angle rather than a quaternion, since the Euler representation boosts the performance of the motion estimation compared to the quaternion representation \cite{zhou2017unsupervised}. A motion vector corresponds to a transformation $\bm{T}$ in $\mathbb{R}^3$, an element in the special Euclidean group SE(3):
\begin{equation}
    \bm{u} \equiv \bm{T} = 
    \begin{bmatrix} 
        \bm{R} & \bm{t} \\
        0 & 1 
    \end{bmatrix},
\end{equation}
where $\bm{R}$ is a rotation matrix in the special orthogonal group SO(3).

The proposed SimVODIS aims to maximize the conditional probability of the motion, $\bm{u}_{t, t-1}$, and depth image, $\bm{D}_t$, given a pair of monocular images $(\bm{I}_{t-1}, \bm{I}_{t})$ at time step $t$ as follows:
\begin{equation}
    (\bm{\hat{u}}_{t, t-1}, \bm{\hat{D}}_{t}) = \underset{(\bm{u}_{t, t-1}, \bm{D}_{t})}{\arg\max} P(\bm{u}_{t, t-1}, \bm{D}_{t}|\bm{I}_{t}, \bm{I}_{t-1}).
\end{equation}
Since SimVODIS performs with two additional separate branches, one for motion estimation and the other for depth map prediction, the probabilities of motion and depth given $(\bm{I}_{t-1}, \bm{I}_{t})$ are independent. Thus, the conditional probability becomes
\begin{align}
    \begin{split}
    P(\bm{u}_{t, t-1}, \bm{D}_{t}|\bm{I}_{t}, \bm{I}_{t-1}) &= P(\bm{u}_{t, t-1}|\bm{I}_{t}, \bm{I}_{t-1}) \cdot P(\bm{D}_{t}|\bm{I}_{t}, \bm{I}_{t-1}) \\
    &= P(\bm{u}_{t, t-1}|\bm{I}_{t}, \bm{I}_{t-1}) \cdot P(\bm{D}_{t}|\bm{I}_{t}).
    \end{split}
\end{align}

In practice, we input a set of three consecutive images when estimating motion vectors because this setting ensures robustness and enhances performance \cite{zhou2017unsupervised, mahjourian2018unsupervised}. With this setting, SimVODIS estimates two motion vectors at a time and the target function of SimVODIS becomes
\begin{equation}
    (\bm{\hat{U}}_{t}, \bm{\hat{D}}_{t}) = \underset{(\bm{U}_{t}, \bm{D}_{t})}{\arg\max} P(\bm{U}_{t}|\bm{I}_{t+1}, \bm{I}_{t}, \bm{I}_{t-1}) \cdot P(\bm{D}_{t}|\bm{I}_{t}),
\end{equation}
where $\bm{U}_{t}=[\bm{u}_{t, t-1}; \bm{u}_{t, t+1}]$.

\subsection{Network Architecture}
\begin{figure*}
    \centering
    \includegraphics[width=0.95\textwidth]{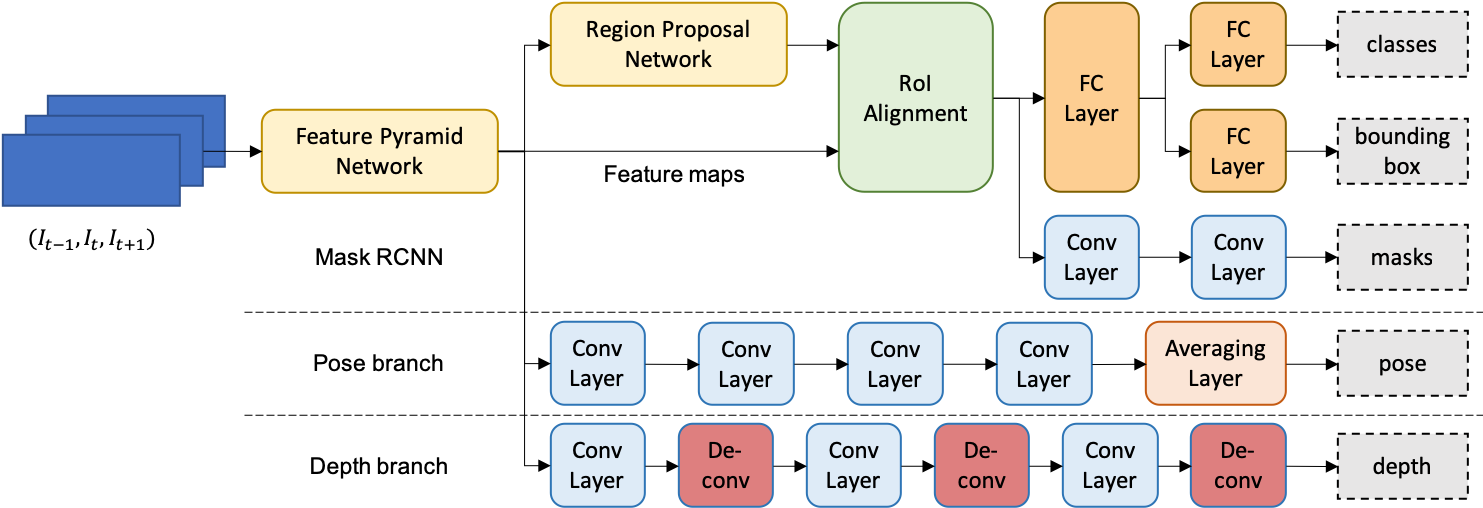}
    \caption{The conceptualized architecture of the proposed SimVODIS network which estimates motion and semantics simultaneously. The network is built on top of the Mask-RCNN architecture by designing pose and depth branches. The pose branch estimates the relative pose between three consecutive image sequences utilizing the features extracted from the feature pyramid network. The depth branch evaluates the depth map of the center image from the extracted features of the center image.}
    \label{fig:network_architecture}
\end{figure*}
Fig. \ref{fig:network_architecture} describes the conceptualized architecture of the proposed SimVODIS network. We design the SimVODIS network based on the following ideas: 1) Mask-RCNN extracts general features for both semantic and geometric tasks such as region proposal, class labeling, bounding box regression and mask extraction and 2) we could use these rich features to estimate the relative pose and predict the depth map since the extracted features are useful for both semantic and geometric tasks. For SimVODIS, we design two network branches: pose and depth branches. The pose branch estimates the relative pose between three consecutive image sequences using the rich features from the feature pyramid network (FPN). Table \ref{tb:network_motion} displays the detailed network architecture of the pose branch. The increase in the total amount of parameters due to the pose branch is minimal.

\begin{figure*}
    \centering
    \includegraphics[width=0.95\textwidth]{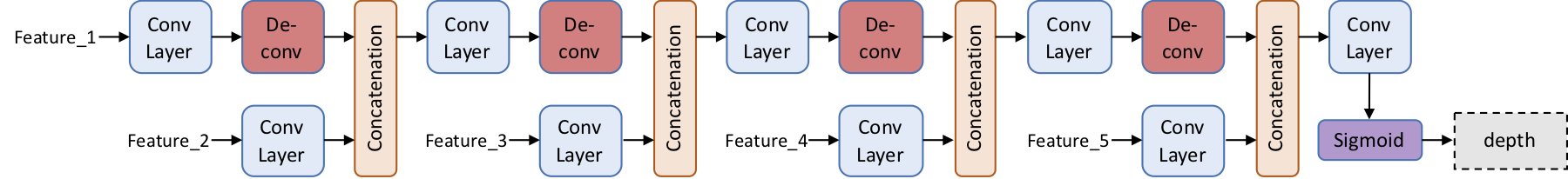}
    \caption{The architecture of the depth branch. It alternates from convolution and de-convolution layers. The final sigmoid layer normalizes the estimated values and produces an inverse depth map.}
    \label{fig:network_depth}
\end{figure*}

Next, Fig. \ref{fig:network_depth} shows the architecture of the depth branch for depth map prediction. The depth branch predicts the inverse depth map rather than directly estimating depth values for numerical stability. The depth branch exploits feature maps at all scales to capture both macroscopic and microscopic characteristics. Previous works predict depth maps in four different scales during training to cope with the gradient locality problem. In contrast, SimVODIS only generates one depth map whose scale equals to the input image, since rich features are already extracted from FPN and the gradient locality problem does not occur for SimVODIS.

\begin{table}[!t]
\renewcommand{\arraystretch}{1.4}
\caption{Configuration of the VO branch}
\label{tb:network_motion}
\centering
\begin{tabular}{c c c c c}
\thickhline
\textbf{Layer} & \textbf{Size} & \textbf{Padding} & \textbf{Stride} & \textbf{Channels}\\
\hline
conv1 & $7 \times 7$ & $0$ & $1$ & $256 \times 3$\\
conv2 & $5 \times 5$ & $0$ & $1$ & $256$\\
conv3 & $3 \times 3$ & $0$ & $1$ & $128$\\
conv4 & $7 \times 7$ & $0$ & $1$ & $64$\\

\thickhline
\end{tabular}
\end{table}

\subsection{Loss Functions}
During the training procedure, the SimVODIS network searches for the optimal hyper-parameters, $(\bm{\theta}_{pose}, \bm{\theta}_{depth})$, that maximize
\begin{equation}
\begin{split}
    (\bm{\theta}^{*}_{pose}, \bm{\theta}^{*}_{depth}) = \underset{(\bm{\theta}_{pose}, \bm{\theta}_{depth})}{\arg\max} [P(\bm{U}_{t}|\bm{I}_{t+1}, \bm{I}_{t}, \bm{I}_{t-1}; \bm{\theta}_{pose}) \cdot \\ P(\bm{D}_{t}|\bm{I}_{t}; \bm{\theta}_{depth})].
\end{split}
\end{equation}
To learn the optimal hyper-parameters, we employ three loss functions to get minimized.

\subsubsection{Image Reconstruction Loss}
The image reconstruction loss takes the fundamental role in training the networks with unlabeled video sequences in an unsupervised manner. We reconstruct the center image ($\bm{I}_{t}$) seen from two nearby camera views using the estimated motion vectors and the predicted depth map of $\bm{I}_t$. For the image reconstruction, we project each pixel in $\bm{I}_t$ onto the nearby views as follows:
\begin{equation}
    p_n = \bm{K} \cdot \bm{\hat{T}}_{t, n} \cdot \bm{\hat{D}}_{t}(p_t) \cdot \bm{K}^{-1} \cdot p_t,
\end{equation}
where $n \in \{t-1, t+1\}$, $\bm{K}$ is the camera intrinsic matrix and $p_t$ represents the homogeneous coordinate of each pixel in $\bm{I}_t$. Since $p_n$ contain continuous values while coordinates of digital images deal with discrete values, we use the differentiable bilinear sampling \cite{jaderberg2015spatial}. Then, the center image is reconstructed as follows:
\begin{equation}
    \bm{\hat{I}}_{recon}(p_t) = \bm{I}_n(p_n) = \sum_{i \in H} \sum_{j \in V} w_{ij}\bm{I}_n(p_{n}^{ij}),
\end{equation}
where $p_{n}^{ij}$ is the 4-pixel neighbors of $p_n$, $w_{ij}$ denotes the spatial proximity between $p_n$ and $p_{n}^{ij}$ and $\sum_{i, j}w_{ij} = 1$, $H = \{left, right\}$, and $V = \{top, bottom\}$. Finally, the reconstruction loss is
\begin{equation}
    L_{recon} = \sum_{p} |\bm{\hat{I}}_{recon, t}(p) - \bm{I}_t(p)|,
\end{equation}
where $p$ is the set of pixel coordinates.

\subsubsection{Structural Similarity Loss}
The image reconstruction loss implicitly assumes that the scenes are Lambertian, thus the brightness remains the same for all observation angles. Violation of this assumption could lead to corrupted gradients, spoiling the training procedure. For improvement of robustness, we use the structural similarity metric defined as
\begin{equation}
    SSIM(x, y) = \frac{(2\mu_x\mu_y + c_1)(2\mu_{xy} + c_2)}{(\mu_{x}^{2} + \mu_{y}^{2} + c_1)(\sigma_x + \sigma_y + c_2)},
\end{equation}
where $x$ and $y$ are image patches, $\mu$ and $\sigma$ are patch means and variances, $c_1=0.01^2$, and $c_2=0.03^2$. We extract $3 \times 3$ image patches from $\bm{I}_t$ and $\bm{I}_{recon, t}$ and minimize
\begin{equation}
    L_{smooth} = \sum_{p} 1 - SSIM(s(\bm{I}_{recon, t}, p), s(\bm{I}_t, p)),
\end{equation}
where $s(\bm{I}, p) = \{\bm{I}(p^x + i, p^y + j)|i,j\in\{-1, 0, 1\} \}$ samples a patch of size $3 \times 3$ centered at $p$ from an image $\bm{I}$.

\subsubsection{Depth Smoothness Loss}
We regularize the depth map estimate to encourage smoothness without which random sharp lines appear in the predicted depth map. By imposing a loss on the gradients of the depth map, we can achieve smoothness over the depth map. In addition, we penalize the smoothness loss by inversely weighting it with the gradients of the input image as follows:
\begin{equation}
    L_{smooth} = \sum|\partial_x \bm{D}_t| \cdot e^{-|\partial_x \bm{I}_x|} + |\partial_y \bm{D}_t| \cdot e^{-|\partial_y \bm{I}_y|}.
\end{equation}
The weighting allows discontinuities for the regions where discontinuities appear in the input image.

\subsection{Training Scheme}
For training SimVODIS, we first freeze the parameters of Mask-RCNN and initialize the parameters of the pose estimation and the depth map prediction branches. Then, we feed a set of three consecutive images at a time. The SimVODIS network takes in the three images as a batch, which keeps the computation time not to increase. The pose branch estimates the camera motion vectors ($\bm{U}_t$). The depth map prediction branch uses only the center image and estimates the depth map. Using the estimated depth map, the depth smoothness loss is evaluated. Finally, the estimated motion vectors and depth map lead to image reconstruction where the reconstruction and SSIM losses are calculated. In the process of calculating the reconstruction loss, we apply the auto-masking technique and the minimum (min) reprojection loss approach \cite{godard2019digging}. Bounding boxes, class labels, and masks of objects in each input image are detected in the process as well.

Conventional methods apply all loss functions at four different scales to overcome the gradient locality. The scale ranges from the input image resolution to $1/8$ of the input resolution ($1$, $1/2$, $1/4$, and $1/8$). However, we do not employ such an approach and use a single image resolution that matches the input resolution. Since the majority of the SimVODIS network is already trained and the pose estimation and the depth map prediction branches are shallow, the problem of gradient locality does not occur.

The final loss function is a weighted sum of the three losses:
\begin{equation}
    L_{final} = \lambda_1 \cdot L_{recon} + \lambda_2 \cdot L_{SSIM} + \lambda_3 \cdot L_{smooth},
\end{equation}
where $\lambda_1$, $\lambda_2$, and $\lambda_3$ are weights. We empirically set $\lambda_1 = 0.15$, $\lambda_2 = 0.85$ and $\lambda_3 = 0.001$.
\section{Evaluation}
We delineate the evaluation settings and methods for the verification of the performance of SimVODIS in this section. Since the weights of the Mask-RCNN part of SimVODIS are frozen and the detection and segmentation performance of SimVODIS matches that of the state-of-the-art, we focus on evaluating the performance of pose estimation and depth map prediction.

\subsection{Datasets}
\begin{figure}
    \centering
    \includegraphics[width=0.45\textwidth]{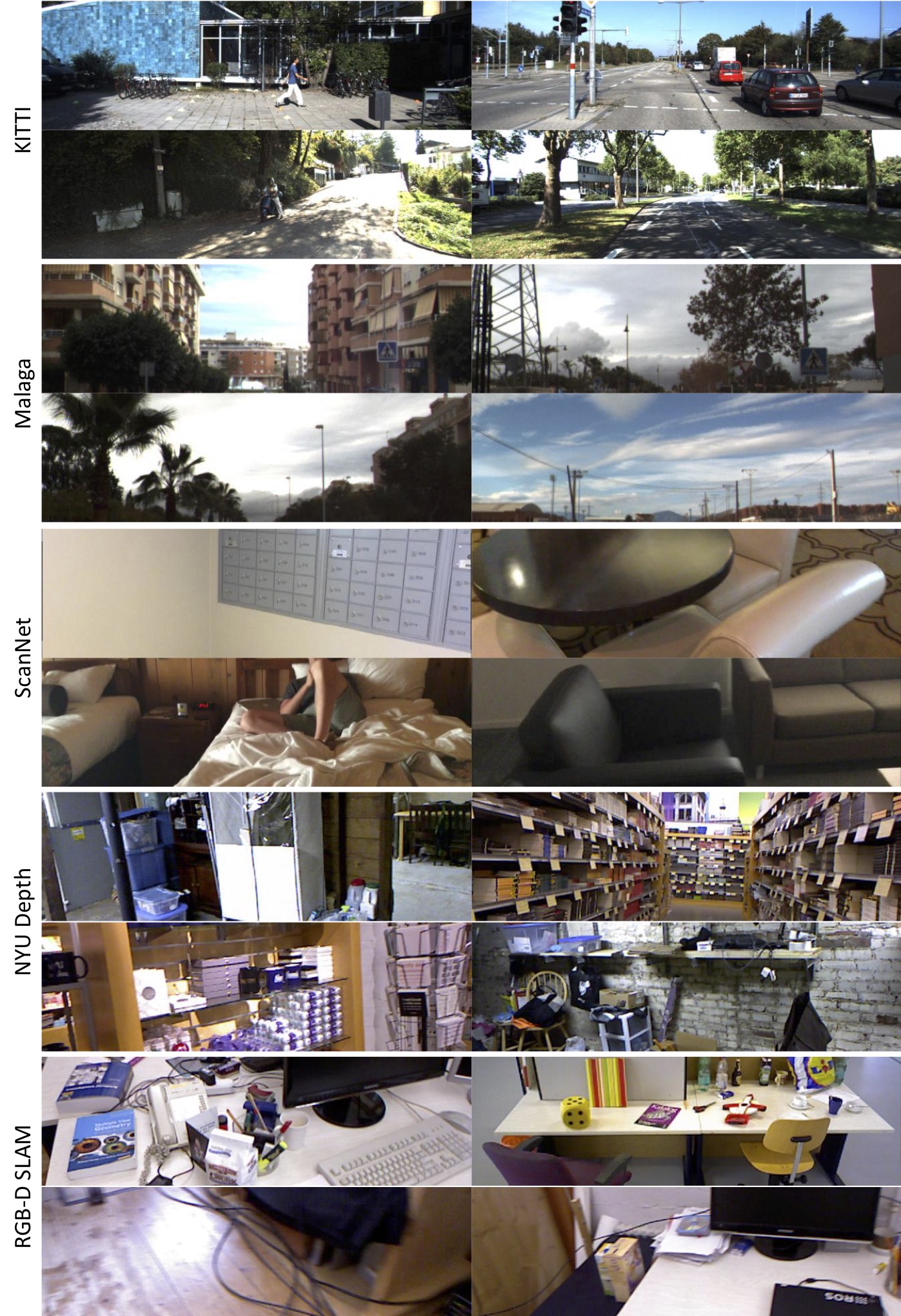}
    \caption{Sample images from the mixture of datasets used for training and testing. The mixture of datasets consists of KITTI, ScanNet, EuRoC, Malaga and NYU depth.}
    \label{fig:dataset_samples}
\end{figure}

Unlike previous works, we use a set of monocular video datasets for training SimVODIS since intelligent agents would work in both outdoor and indoor environments. In total, our study includes seven datasets: five for mainly training (main datasets) and two for additional testing (extra datasets). Fig. \ref{fig:dataset_samples} shows sample images from the main datasets and Table \ref{tb:datasets_summary} summarizes the characteristics of each main dataset. First of all, we train SimVODIS on the KITTI dataset following the convention \cite{godard2019digging}. Then, we compare the performance of pose estimation and depth map prediction against baselines using the standard KITTI split \cite{eigen2014depth}. Next, we train SimVODIS on the combinations of main datasets. In total, we generate five versions of SimVODIS and test them on various datasets. This allows the examination of the effect of the dataset heterogeneity on the performance of joint estimation of ego-motion and depth map prediction.

\begin{table*}[!t]
\renewcommand{\arraystretch}{1.4}
\caption{Summary of Datasets for Training}
\label{tb:datasets_summary}
\centering
\begin{tabular}{l | c c c c c c}
\thickhline
\multicolumn{1}{c|}{\textbf{Dataset}} & \textbf{Publisher} & \textbf{Modality} & \textbf{FPS} & \textbf{Resolution} & \textbf{Sequences} & \textbf{Environment}\\
\hline
KITTI \cite{geiger2013vision}& KIT & Stereo & 10Hz & 1241$\times$376 & 21 & Outdoor\\
M{\'a}laga \cite{blanco2014malaga}& MRPT & Stereo & 2$\times$20Hz & 1024$\times$768 or 800$\times$600 & 15 & Outdoor\\
ScanNet \cite{dai2017scannet}& Stanford & RGB-D & 15 or 30Hz & 1296$\times$968 & 1513 & Indoor\\
NYU Depth \cite{silberman2012indoor}& NYU & RGB-D & 20-30Hz & 640$\times$480 & 464 & Indoor\\
RGB-D SLAM \cite{sturm2012benchmark}& TUM & RGB-D & 30Hz & 640$\times$480 & 39 & Indoor\\

\thickhline
\end{tabular}
\end{table*}

\subsubsection{Description of Main Datasets}
We employ the following five datasets as main datasets for training SimVODIS and verification of performance.
\begin{itemize}
    \item KITTI \cite{geiger2013vision}: The KITTI benchmark is one of the most well-known public datasets for evaluating the performance of VO and visual SLAM algorithms. It was collected during an outdoor car driving scenario using a stereo camera. Since a number of dynamic objects appear in the scenes and the car moves fast, the KITTI benchmark poses a challenge for VO and visual SLAM algorithms. It consists of 22 sequences and 11 of them (sequence 0-10) provide ground-truth camera motion.
    \item M{\'a}laga \cite{blanco2014malaga}: Similar to the KITTI benchmark, the M{\'a}laga dataset was collected during an urban car driving scenario using a stereo camera. It consists of 15 sequences whose ground-truth camera motion is not available. Instead, it offers other sensor data such as Lidar and GPS.
    \item ScanNet \cite{dai2017scannet}: The ScanNet dataset encompasses 1,513 indoor RGB-D sequences. The dataset offers rich information regarding the scenes including camera motion, surface reconstruction, and semantic annotation. Thus, a number of vision tasks can utilize the dataset for training and evaluation.
    \item NYU depth \cite{silberman2012indoor}: The NYU depth dataset (Version 2) includes 464 RGB-D sequences. It targets indoor scene understanding and it does not provide ground-truth camera motion data. The dataset deals with various types of indoor spaces such as kitchen, office room, living room, etc.
    \item RGB-D SLAM \cite{sturm2012benchmark}: The RGB-D SLAM dataset comprises 39 indoor RGB-D sequences. Its original purpose is for the development and evaluation of VO and SLAM systems. It provides ground-truth camera pose data measured by a high-quality motion capture system. The camera was held by either humans or robots during the data collection process.
\end{itemize}

\subsubsection{Description of Extra Datasets}
We adopt two additional datasets to test the performance of SimVODIS in unseen environments.
\begin{itemize}
    \item Make3D \cite{saxena2008make3d}: The Make3D dataset provides 134 pairs of an outdoor image and the corresponding depth map. The images are not consecutive but capture separate views. The resolutions of the images and the depth maps are 1,704$\times$2,272 and 55$\times$305, respectively and the depth maps are acquired with a laser device.
    \item 7 Scenes \cite{shotton2013scene}: As the name implies, the dataset consists of seven indoor sequences. Each sequence contains a record of a continuous stream of tracked RGB-D camera frames. The lengths of sequences range from 500 to 1,000. Both images and depth maps have the resolution of 640$\times$480.
\end{itemize}

\subsubsection{Combination of Datasets for Training and Testing}
\begin{table}[!t]
\renewcommand{\arraystretch}{1.4}
\caption{SimVODIS Variants and Corresponding Datasets}
\label{tb:variants}
\centering
\begin{tabular}{l | c | c | c | c | c}
\thickhline
\multicolumn{1}{c|}{\textbf{Model Name}} & \textbf{KITTI} & \textbf{M{\'a}laga} & \textbf{ScanNet} & \textbf{NYU} & \textbf{RGB-D}\\
\hline
SimVODIS$_{k}$ & \checkmark & - & - & - & -\\
SimVODIS$_{o}$ & \checkmark & \checkmark & - & - & -\\
SimVODIS$_{r}$ & - & - & - & - & \checkmark\\
SimVODIS$_{i}$ & - & - & \checkmark & \checkmark & \checkmark\\
SimVODIS$_{a}$ & \checkmark & \checkmark & \checkmark & \checkmark & \checkmark\\
\thickhline
\end{tabular}
\end{table}

We combine the main datasets in five ways and generate five variants of SimVODIS. Table \ref{tb:variants} describes how the datasets are combined and the subscriptions in the table represent KITTI (k), outdoor (o), RGB-D (r), indoor (i), and all (a), respectively. First of all, we use the KITTI dataset for performance verification (SimVODIS$_{k}$ corresponds to SimVODIS$_{full}$) compared to baselines. For this, we split the dataset into train, validation and test sets as conventional methods \cite{zhou2017unsupervised}. Then, we use sequences 9 and 10 to demonstrate the pose estimation performance and 697 images selected to verify the depth map prediction performance.

Next, we examine the effect of dataset heterogeneity on the model performance by combining the datasets as described in Table \ref{tb:variants}. We analyze the model performance on both KITTI (outdoor) and RGB-D SLAM (outdoor) test sets. We select a part of image frames from the RGB-D SLAM dataset on which the models are not trained to assess the performance in indoor environments. In addition, we use the extra datasets for testing the performance. Since we do not train SimVODIS on the extra datasets, testing SimVODIS models on the extra datasets reveals the generalization performance according to the dataset heterogeneity of training data.

\subsection{Metrics}
For intuitive comparison of the proposed SimVODIS against baselines, we employ both quantitative and qualitative methods. We describe the metrics of the quantitative analysis for clarity in this subsection.

\subsubsection{Pose Estimation}
We quantitatively evaluate the performance of the pose estimation by Absolute Trajectory Error (ATE) \cite{sturm2012benchmark}. For convenience, we briefly summarize ATE as follows:
\begin{equation}
    ATE = (\frac{1}{N} \sum_{i}||\Delta \textbf{p}_{i}||^2 )^{\frac{1}{2}},
\end{equation}
where $N$ is the number of estimated states and $\Delta \textbf{p}_{i}$ is the position difference (error) between the estimated state and the ground-truth at the $i$-th step.

\subsubsection{Depth Map Prediction}
The following metrics represent quantitative performance of depth map prediction: Absolute Relative Difference (ARD) \cite{saxena2008make3d}, Squared Relative Difference (SRD), Root Mean Square Error (RMSE) \cite{li2010towards}, RMSE log \cite{eigen2014depth} and three classes of Thresholds ($\delta < \nu$, $\nu \in \{1.25, 1.25^2, 1.25^3\}$) \cite{ladicky2014pulling}. By letting $\bm{D}$ and $\hat{\bm{D}}$ denote the ground-truth depth map and the predicted depth map, respectively, we calculate each metric as follows:
\begin{equation}
\text{ARD} = \frac{1}{N(p)} \sum_{p} \frac{|\bm{D}(p) - \hat{\bm{D}}(p)|}{\bm{D}(p)}
\end{equation}

\begin{equation}
\text{SRD} = \frac{1}{N(p)} \sum_{p} \frac{|\bm{D}(p) - \hat{\bm{D}}(p)|^2}{\bm{D}(p)}
\end{equation}

\begin{equation}
\text{RMSE} = \sqrt{\frac{1}{N(p)} \sum_{p} |\bm{D}(p) - \hat{\bm{D}}(p)|^2}
\end{equation}

\begin{equation}
\text{RMSE}_{log} = \sqrt{\frac{1}{N(p)} \sum_{p} |\log\bm{D}(p) - \log\hat{\bm{D}}(p)|^2}
\end{equation}

\begin{equation}
\delta = \max(\frac{\bm{D}(p)}{\hat{\bm{D}}(p)}, \frac{\hat{\bm{D}}(p)}{\bm{D}(p)}) < \nu
\end{equation}

\subsection{Ablation Study}
To investigate the effect of each design choice of SimVODIS, we train SimVODIS in multiple ways. We control the following components of SimVODIS: auto-masking, min loss, SSIM loss, number of depth map scales during training (either $1$ or $4$), and input image size (either $416 \times 128$ or $640 \times 192$). For the full SimVODIS model, we apply all the auto-masking technique, the min loss, and the SSIM loss and use one depth map scale and the input image of size $640 \times 192$. In total, we train five variants of SimVODIS and each variant differs one training condition from the full SimVODIS model (SimVODIS$_{k}$).

\subsection{Baselines}

\begin{table}[!t]
\renewcommand{\arraystretch}{1.4}
\caption{Baselines and Their Functionalities}
\label{tb:baselines}
\centering
\begin{tabular}{l | c | c | c}
\thickhline
\multicolumn{1}{c|}{\textbf{Method}} & \textbf{Pose} & \textbf{Depth} & \textbf{Features}\\
\hline
Zhou et al. \cite{zhou2017unsupervised} & \checkmark & \checkmark & Unsupervised\\
Mahjourian et al. \cite{mahjourian2018unsupervised} & \checkmark & \checkmark & Unsupervised\\
Ranjan et al. \cite{ranjan2019competitive} & \checkmark & \checkmark & Unsupervised\\
Casser et al. \cite{casser2019depth} & \checkmark & \checkmark & Unsupervised\\
Godard et al. \cite{godard2019digging} & - & \checkmark & Unsupervised\\
\hline
ORB-SLAM \cite{mur2017orb} & \checkmark & - & Hand-crafted\\
\hline
Eigen et al. \cite{eigen2014depth} & - & \checkmark & Supervised\\
Liu et al. \cite{liu2015learning} & - & \checkmark & Supervised\\
Fu et al. \cite{fu2018deep} & - & \checkmark & Supervised\\
\hline
\textbf{SimVODIS} & \checkmark & \checkmark & Unsupervised\\

\thickhline
\end{tabular}
\end{table}

Table \ref{tb:baselines} lists the baselines that we compare with SimVODIS. A part of the baselines performs both pose estimation and depth map prediction, while the other part of the baselines conducts either pose estimation or depth map prediction. All the baselines are based on deep feature learning except ORB-SLAM \cite{mur2017orb} which uses hand-crafted features. The seven out of ten deep feature learning methods take unsupervised learning settings, while the other requires depth supervision for training depth map prediction networks.


\section{Results and Analysis}
In this section, we present the evaluation results of SimVODIS in a number of conditions, analyze the results in a thorough manner and establish the effectiveness of SimVODIS.

\subsection{Ablation Study}
\begin{table*}[!t]
\renewcommand{\arraystretch}{1.4}
\caption{Quantitative Depth Evaluation Results According to Various Training Conditions}
\label{tb:result_depth_quantitative_ablation}
\centering
\begin{tabular}{l | c || c c c c | c c c}
\thickhline
\multicolumn{1}{c|}{\textbf{Training Condition}} & \textbf{Cap} & \textbf{ARD} & \textbf{SRD} & \textbf{RMSE} & \textbf{RMSE$_{log}$} & \textbf{$\delta < 1.25$} & \textbf{$\delta < 1.25^2$} & \textbf{$\delta < 1.25^3$}\\
\hline
SimVODIS w/o automasking & 80m & 0.129  &   0.865  &   4.905  &   0.199  &   0.844  &   0.957  &   0.983\\
SimVODIS w/o min. loss & 80m & 0.127  &   0.853  &   4.876  &   0.199  &   0.846  &   0.955  &   0.982\\
SimVODIS w/o SSIM loss & 80m & 0.136  &   0.876  &   4.997  &   0.205  &   0.832  &   0.955  &   0.983\\
SimVODIS with $N(scales) = 4$ & 80m & 0.126  &   0.824  &   4.823  &   0.196  &   0.847  &   0.958  &   \textbf{0.984}\\
SimVODIS with 416$\times$128 & 80m & 0.128  &   0.852  &   4.976  &   0.200  &   0.844  &   0.956  &   0.983\\
\hline
\textbf{SimVODIS$_{full}$} & 80m & \textbf{0.123}  & \textbf{0.797}  & \textbf{4.727}  &   \textbf{0.193}  &   \textbf{0.854}  &  \textbf{0.960}  &  \textbf{0.984}\\
\thickhline
\end{tabular}
\end{table*}

\begin{table}[!t]
\renewcommand{\arraystretch}{1.4}
\caption{Quantitative Pose Estimation Evaluation Result According to Various Training Conditions}
\label{tb:result_pose_quantitative_ablation}
\centering
\begin{tabular}{l | c | c}
\thickhline
\multicolumn{1}{c|}{\textbf{Method}} & \textbf{Seq. 09} & \textbf{Seq. 10}\\
\hline
SimVODIS w/o automasking & 0.019 $\pm$ 0.010 & 0.013 $\pm$ 0.008\\
SimVODIS w/o min. loss & 0.014 $\pm$ 0.008 & 0.012 $\pm$ 0.008\\
SimVODIS w/o SSIM loss & 0.017 $\pm$ 0.010 & 0.014 $\pm$ 0.011\\
SimVODIS with $N(scales) = 4$ & 0.014 $\pm$ 0.007 & 0.013 $\pm$ 0.008\\
SimVODIS with 416$\times$128 & 0.014 $\pm$ 0.007 & \textbf{0.011} $\pm$ \textbf{0.008}\\
\hline
\textbf{SimVODIS$_{full}$} & \textbf{0.012} $\pm$ \textbf{0.006} & \textbf{0.011} $\pm$ \textbf{0.008}\\

\thickhline
\end{tabular}
\end{table}

Tables \ref{tb:result_depth_quantitative_ablation} and \ref{tb:result_pose_quantitative_ablation} describe the results of ablation study. First of all, the automasking technique, the min loss, and the SSIM loss all contribute to clean gradient for training the networks. Missing one of the techniques degrades the performance of both depth map prediction and pose estimation. The gradient gets deteriorated by moving objects, occluded views and illumination variation without the automasking technique, the min. loss and the SSIM loss, respectively. Among the three remedies for the corrupted gradient, the effect of the SSIM loss is overwhelming as the performance degradation is the greatest without the SSIM loss. From this, we could infer that the illumination variation is more common than moving objects and occluded views.

Next, comparing the performance of SimVODIS$_{full}$ and SimVODIS with $N(scales)=4$ reveals that SimVODIS does not entail the problem of gradient locality. Employing pretrained Mask-RCNN allows the design of shallower networks for depth map prediction and pose estimation and using rich features from FPN accelerates the training procedure. The performance of SimVODIS with $N(scales)=4$ almost matches that of SimVODIS$_{full}$ in depth map prediction though the performance of SimVODIS with $N(scales)=4$ in pose estimation slightly drops.

The last two rows of Tables \ref{tb:result_depth_quantitative_ablation} and \ref{tb:result_pose_quantitative_ablation} indicate that using smaller images marginally reduces the performance of both depth map prediction and pose estimation. We assume that the networks could receive more context as the input images get larger and this helps improve the performance. This tendency corresponds to the result reported in the previous work \cite{godard2019digging}.

\subsection{Comparison Against Baselines}
\begin{table*}[!t]
\renewcommand{\arraystretch}{1.4}
\caption{Quantitative Depth Evaluation Results Compared to Baselines}
\label{tb:result_depth_quantitative}
\centering
\begin{tabular}{l | c c || c c c c | c c c}
\thickhline
\multicolumn{1}{c|}{\textbf{Method}} & \textbf{Supervision} & \textbf{Cap} & \textbf{ARD} & \textbf{SRD} & \textbf{RMSE} & \textbf{RMSE$_{log}$} & \textbf{$\delta < 1.25$} & \textbf{$\delta < 1.25^2$} & \textbf{$\delta < 1.25^3$}\\
\hline
Eigen et al. \cite{eigen2014depth} Coarse & Depth & 80m & 0.214 & 1.605 & 6.563 & 0.292 & 0.673 & 0.884 & 0.957\\
Eigen et al. \cite{eigen2014depth} Fine & Depth & 80m & 0.203 & 1.548 & 6.307 & 0.282 & 0.702 & 0.890 & 0.958\\
Liu et al. \cite{liu2015learning} & Depth & 80m & 0.201 & 1.584 & 6.471 & 0.273 & 0.680 & 0.898 & 0.967\\
Fu et al. \cite{fu2018deep} & Depth & 80m & 0.072 & 0.307 & 2.727 & 0.120 & 0.932 & 0.984 & 0.994\\
\hline
Zhou et al. \cite{zhou2017unsupervised} & - & 80m & 0.208 & 1.768 & 6.856 & 0.283 & 0.678 & 0.885 & 0.957\\
Mahjourian et al. \cite{mahjourian2018unsupervised} & - & 80m & 0.163 & 1.240 & 6.220 & 0.250 & 0.762 & 0.916 & 0.968\\
Ranjan et al. \cite{ranjan2019competitive} & - & 80m & 0.148 & 1.149 & 5.464 & 0.226 & 0.815 & 0.935 & 0.973\\
Casser et al. \cite{casser2019depth} & - & 80m & 0.141 & 1.026 & 5.291 & 0.215 & 0.816 & 0.945 & 0.979\\
Godard et al. \cite{godard2019digging} & - & 80m & \textbf{0.115} & 0.903 & 4.863 & \textbf{0.193} & \textbf{0.877} & 0.959 & 0.981\\
\textbf{SimVODIS} & - & 80m & 0.123  & \textbf{0.797}  & \textbf{4.727}  &   \textbf{0.193}  &   0.854  &  \textbf{0.960}  &  \textbf{0.984}\\
\thickhline
\end{tabular}
\end{table*}

\begin{table}[!t]
\renewcommand{\arraystretch}{1.4}
\caption{Quantitative Evaluation Result of Pose Estimation Compared to Baselines}
\label{tb:result_pose_quantitative}
\centering
\begin{tabular}{l | c | c}
\thickhline
\multicolumn{1}{c|}{\textbf{Method}} & \textbf{Seq. 09} & \textbf{Seq. 10}\\
\hline
ORB-SLAM (full) \cite{mur2017orb} & 0.014 $\pm$ 0.008 & 0.012 $\pm$ 0.011\\
ORB-SLAM (short) \cite{mur2017orb} & 0.064 $\pm$ 0.141 & 0.064 $\pm$ 0.130\\
\hline
Mean Odom. & 0.032 $\pm$ 0.026 & 0.028 $\pm$ 0.023\\
Zhou et al. (5-frame)\cite{zhou2017unsupervised} & 0.021 $\pm$ 0.017 & 0.020 $\pm$ 0.015\\
Mahjourian et al. w/o ICP \cite{mahjourian2018unsupervised} & 0.014 $\pm$ 0.010 & 0.013 $\pm$ 0.011\\
Mahjourian et al. with ICP \cite{mahjourian2018unsupervised} & 0.013 $\pm$ 0.010 & 0.013 $\pm$ 0.011\\
Ranjan et al. \cite{ranjan2019competitive} & 0.012 $\pm$ 0.007 & 0.012 $\pm$ \textbf{0.008}\\
Casser et al. \cite{casser2019depth} & \textbf{0.011} $\pm$ \textbf{0.006} & \textbf{0.011} $\pm$ 0.010\\
\hline
\textbf{SimVODIS} & 0.012 $\pm$ \textbf{0.006} & \textbf{0.011} $\pm$ \textbf{0.008}\\

\thickhline
\end{tabular}
\end{table}


Tables \ref{tb:result_depth_quantitative} and \ref{tb:result_pose_quantitative} depict the quantitative performance of SimVODIS compared to baseline models. SimVODIS outperforms or performs on par with baselines in both depth map prediction and pose estimation. Although the depth supervised method \cite{fu2018deep} achieves the performance of the state-of-the-art, five out of seven metrics indicate that SimVODIS outperforms unsupervised depth map prediction methods. In addition, SimVODIS could perform both depth map prediction and pose estimation, while the depth supervised method cannot. For pose estimation, the performance of SimVODIS outperforms or matches baselines. Since the scenes of the KITTI dataset contain multiple dynamic objects, the performance of pose estimation would improve by tackling dynamic objects.

\begin{figure*}
    \centering
    \includegraphics[width=0.98\textwidth]{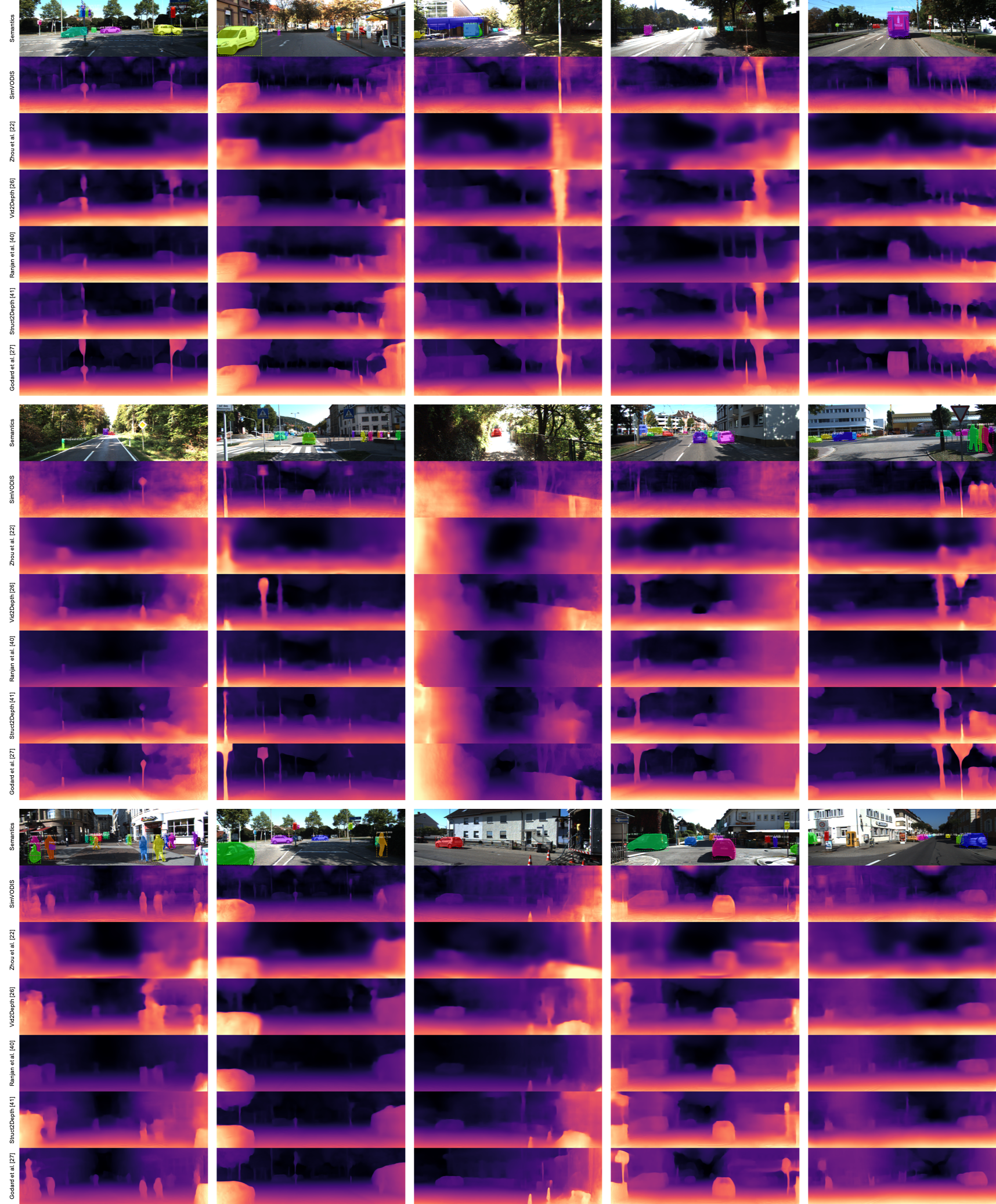}
    \caption{Qualitative comparison of single-view depth map prediction. SimVODIS extracts fine-grained depth maps compared to baselines. The contours of objects are clearly seen in the depth maps from SimVODIS. On the other hand, the contours of objects get crumbled in depths maps from baselines.}
    \label{fig:result_depth_qualitative}
\end{figure*}

Fig. \ref{fig:result_depth_qualitative} presents the qualitative depth map prediction result of SimVODIS and baselines for intuitive understanding. The qualitative results imply that SimVODIS extracts fine-grained depth maps and recovers the depth of objects with higher accuracy compared to baselines. The depth maps from SimVODIS include the outlines of objects and objects can be recognized even from depth maps. In contrast, the outlines of objects get crumbled in depth maps from baselines and they cannot be recognized.

\begin{figure*}
    \centering
    \includegraphics[width=0.98\textwidth]{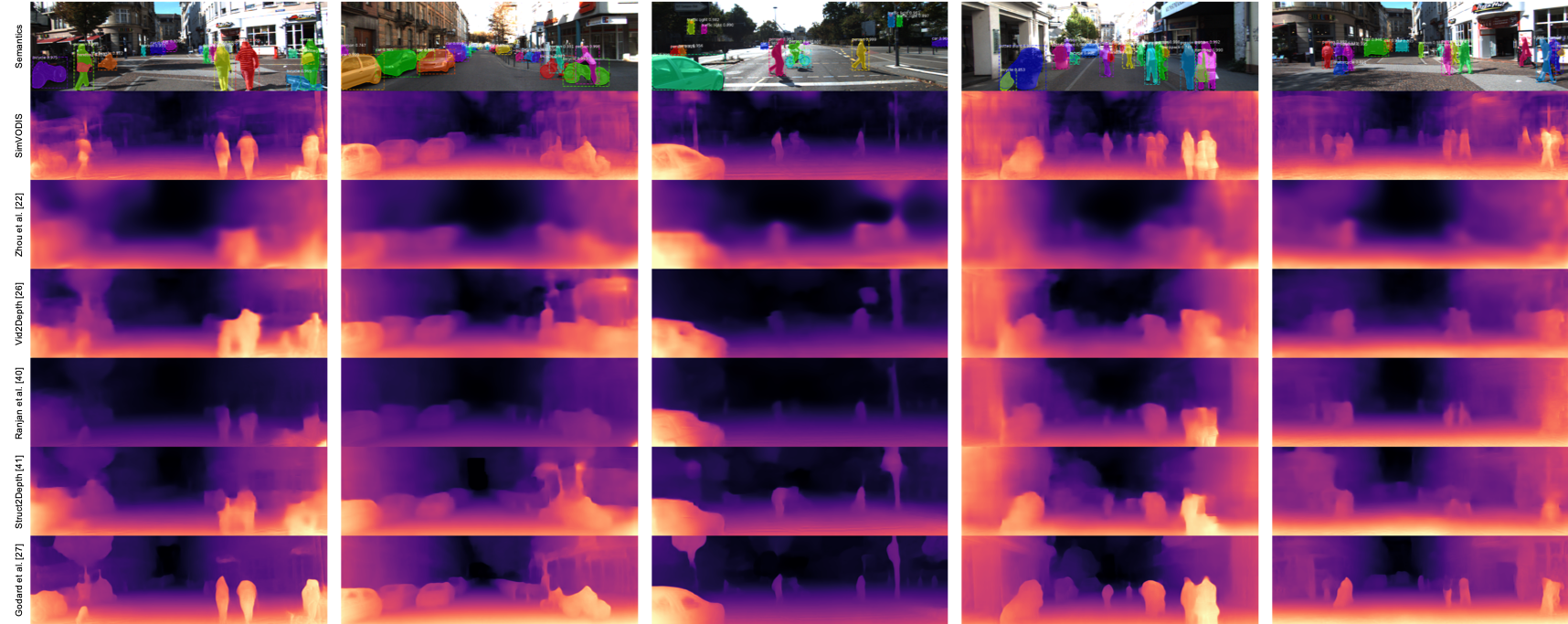}
    \caption{Single-view depth map prediction for scenes containing people and moving objects. SimVODIS captures the outlines of people and moving objects and they can be recognized even in depth maps. On the other hand, baselines cannot recover the depth information of people and moving objects.}
    \label{fig:result_depth_qualitative_people}
\end{figure*}

Fig. \ref{fig:result_depth_qualitative_people} displays the depth map prediction results for the scenes containing people and moving objects. Although people and moving objects in the scenes hinder the prediction of depth maps for baselines, SimVODIS could recover the depths of people and moving objects. In most cases, one can even count the number of people presented in the scenes only looking at the predicted depth maps. We conjecture that the features from FPN contain the information of objects and these features help deal with moving objects for depth map prediction.

\begin{figure*}
    \centering
    \includegraphics[width=0.98\textwidth]{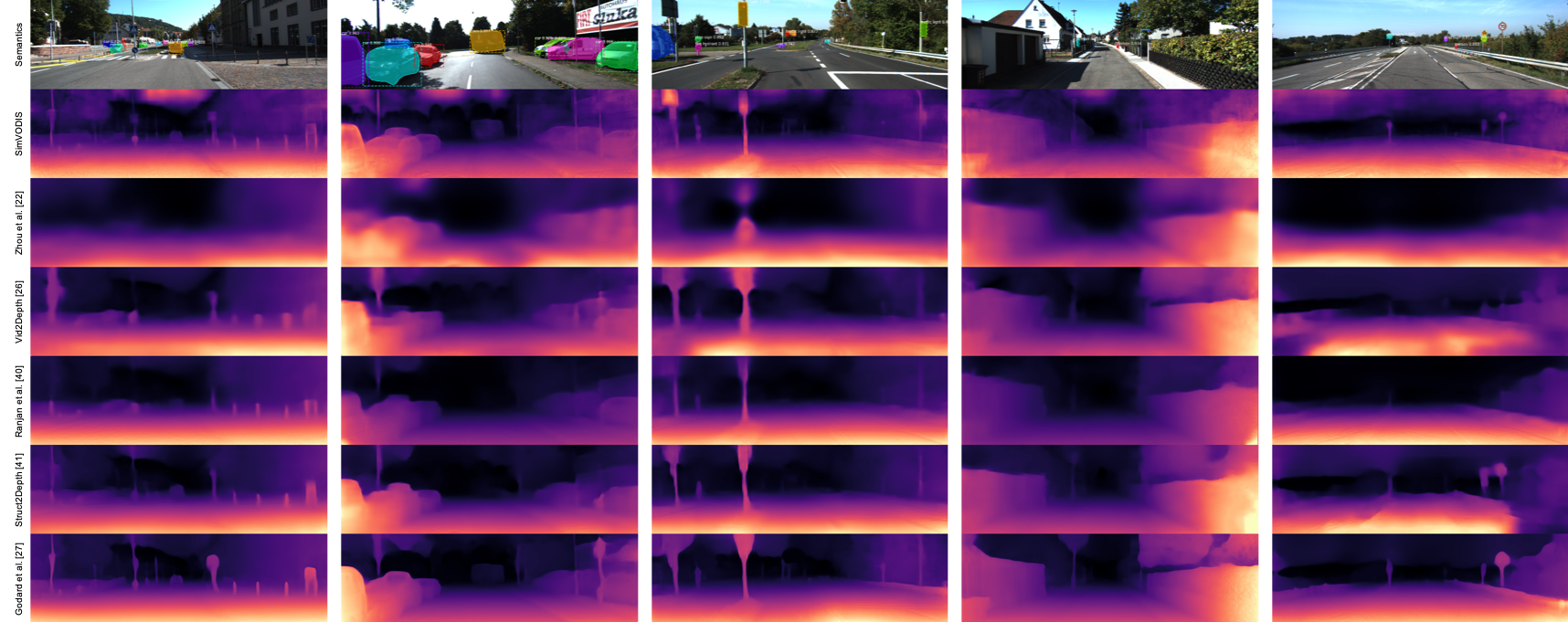}
    \caption{Failure cases of single-view depth map prediction. SimVODIS do not recover the depth of the sky from time to time.}
    \label{fig:result_depth_qualitative_sky}
\end{figure*}

Fig. \ref{fig:result_depth_qualitative_sky} illustrates failure cases of SimVODIS. SimVODIS occasionally predicts small values for the depths of the sky. This phenomenon is not consistent but occurs infrequently. Thus, SimVODIS produces reasonably large values for the depths of the sky for other cases. Though the depths of the sky are not precise in a few scenes, SimVODIS could consistently recover the boundaries of objects when predicting depth maps.

\subsection{Effect of Dataset Heterogeneity}

\begin{table*}[!t]
\renewcommand{\arraystretch}{1.4}
\caption{SimVODIS Variants and Corresponding Datasets}
\label{tb:heterogeneity_depth}
\centering
\begin{tabular}{c | c | c | c | c | c | c | c | c | c | c | c | c}
\thickhline
\multirow{2}{*}{{\textbf{Model}}} & \multicolumn{3}{c|}{\textbf{KITTI} (outdoor)} & \multicolumn{3}{c|}{\textbf{Make3D} (outdoor)} & \multicolumn{3}{c|}{\textbf{RGB-D} (indoor)} & \multicolumn{3}{c}{\textbf{7 Scenes} (indoor)}\\
\cline{2-13}
& ARD & SRD & RMSE & ARD & SRD & RMSE & ARD & SRD & RMSE & ARD & SRD & RMSE \\
\hline
SimVODIS$_{k}$ & \textbf{0.123} & \textbf{0.797} & \textbf{4.727} & 0.392 & 5.407 & 8.958 & 0.322 & 2.971 & 5.797 & 0.253 & 0.249 & 0.682 \\
SimVODIS$_{o}$ & 0.131 & 0.823 & 4.850 & 0.383 & 4.913 & 8.615 & 0.335 & 2.624 & 5.813 & 0.340 & 0.484 & 0.936 \\
SimVODIS$_{r}$ & -                         & - & - & - & - & - & \textbf{0.285} & \textbf{2.302} & \textbf{5.632} & \textbf{0.196} & \textbf{0.119} & \textbf{0.454} \\
SimVODIS$_{i}$ & -                         & - & - & - & - & - & 0.297 & 2.391 & 5.765 & 0.215 & 0.143 & 0.496 \\
SimVODIS$_{a}$ & 0.203 & 1.451 & 7.001 & \textbf{0.367} & \textbf{4.267} & \textbf{8.388} & 0.339 & 2.675 & 6.046 & 0.296 & 0.362 & 0.783 \\
\thickhline
\end{tabular}
\end{table*}

Although it is straightforward that testing on a different dataset from the training dataset would inevitably reduce the performance, no study has investigated how much the performance drop is in the context of joint estimation of ego-motion and depth map. In addition, it is crucial to guarantee the generalization of trained models for robust performance. We aim to investigate the effect of dataset heterogeneity in this section.

We focus on depth map prediction when analyzing the effect of dataset heterogeneity for two reasons. First, each sequence imposes a different level of difficulty for pose estimation, thus pose estimation does not provide absolute-scale performance metrics. Second, the performance of depth map prediction and pose estimation shows a similar tendency as illustrated in the ablation study and the comparative study.

Table \ref{tb:heterogeneity_depth} summarizes the effect of dataset heterogeneity on the performance. Since indoor sequences only provide scenes with depth values less than 15m, we did not test SimVODIS$_{r}$ and SimVODIS$_{i}$, which were trained on indoor scenes, on outdoor sequences. Comparing model performance on KITTI and Make3D indicates that increased heterogeneity of training dataset tends to reduce overfitting in the case of outdoor scenes. The order of model performance is inversely proportional to the number of datasets used for training in the case of KITTI, while the order is directly proportional in the case of Make3D.

However, the trend is not obvious for indoor scenes. Although SimVODIS$_{r}$ and SimVODIS$_{i}$ performs the best for RGB-D as expected, the performance order of SimVODIS$_{k}$, SimVODIS$_{o}$ and SimVODIS$_{a}$ is maintained when tested on KITTI and when tested on RGB-D. Moreover, the performance order is the same for RGB-D and 7 Scenes. We presume that the statistical correspondence between color images and depth maps is similar among KITTI, RGB-D and 7 Scenes. Plus, the low RMSE of 7 Scenes suggests that the average depth value of 7 Scenes is lower than that of RGB-D.
\section{Discussion}
SimVODIS concurrently estimates camera motion vectors and semantics (bounding boxes, class labels and object masks), and predicts depth maps in one thread rather than two threads. SimVODIS provides an unsupervised learning framework with which it learns the optimal parameters from unlabeled video sequences. SimVODIS shows state-of-the-art or comparable performance in motion estimation and depth map prediction. However, there still exist a few future works for further improvement of the performance.

First of all, we could extend SimVODIS to a SLAM system. SimVODIS similar to other VO systems accumulates estimation errors over the course of camera movement. Integrating a loop-closure algorithm offsets the accumulated errors and the system could robustly perform for a longer sequence of observations. Moreover, the SLAM system with SimVODIS could possibly improve the performance of the loop closure detection by utilizing the rich semantics extracted in addition to the conventional geometric features. Current SLAM methods, in general, rely on a visual bag of words for the loop closure detection \cite{mur2017orb}.

Next, we could investigate the application of SimVODIS in dynamic environments. Although SimVODIS itself allows intelligent agents to perform various tasks, the performance gets deteriorated when moving entities appear in the environment. A number of SLAM approaches for dynamic environments employ semantic SLAM in addition to hand-crafted features to handle moving entities \cite{kim2016effective, bescos2018dynaslam}. We expect SimVODIS would show satisfactory performance in dynamic environments with a few add-ons since SimVODIS already extracts semantic information as well as physical information (ego-motion and depth map) in an efficient manner.

In addition, we could improve the accuracy of SimVODIS by enforcing consistency over multiple image frames. Multiple observations of the same entity from multiple views could compensate for recognition errors. Such methods include bundle adjustment \cite{triggs1999bundle} and consistency over multi-views \cite{jeong2017object}. In the following research, we could design a computational method that utilizes multiple observations to enhance the performance of SimVODIS. Moreover, we would consider batch-based computation and parallel processing between GPU and CPU to ensure computational efficiency while enhancing the performance.

Furthermore, we could devise a training scheme for employing multiple training datasets to secure the robust performance of SimVODIS in both indoor and outdoor environments. The investigation of the effect of dataset heterogeneity in our study implies that the current training approach for data-driven VO would not guarantee versatile performance when the environment varies. We plan to collect a number of datasets for training SimVODIS and design a training method for effective feature extraction from a set of training datasets.

Last but not least, we could enhance the architecture of SimVODIS to support different types of imaging modalities such as rgb-d and stereo cameras. The current implementation of SimVODIS supports monocular image sequences that provide less amount of information compared to rgb-d and stereo cameras. We would investigate ways to extract and exploit the extra information inherent in those imaging modalities while keeping the support for monocular cameras. This would solve the scale ambiguity problem and greatly improve the accuracy of pose estimation as well as depth map prediction. Moreover, we expect the data-driven VO following this development pathway will replace the conventional feature-based VO/SLAM in the end with better performance.
\section{Conclusion}
We proposed SimVODIS which concurrently estimates both semantics and physical information inherent in environments when receiving a set of monocular image frames. SimVODIS is the first fully data-driven semantic VO. In contrast to conventional approaches where semantics, poses and depth maps are evaluated in separate modules, SimVODIS extracts all the information in one thread. SimVODIS achieves the state-of-the-art performance in both depth map prediction and pose estimation and its performance in object detection and instance segmentation matches that of state-of-the-art. The depth maps from SimVODIS clearly depict object boundaries which were mashed in conventional methods. Moreover, we examined the effect of dataset heterogeneity on model performance and identified the future research direction to guarantee the generalization of trained models. We expect intelligent agents could realize practical services for humans by deeply understanding surrounding environments with SimVODIS.

\ifCLASSOPTIONcompsoc
  \section*{Acknowledgments}
\else
  \section*{Acknowledgment}
\fi

This work was supported by Institute for Information \& communications Technology Promotion (IITP) grant funded by the Korea government (MSIT) (No. 2018-0-00677, Development of Robot Hand Manipulation Intelligence to Learn Methods and Procedures for Handling Various Objects with Tactile Robot Hands)

\ifCLASSOPTIONcaptionsoff
  \newpage
\fi



\bibliographystyle{IEEEtran}
\bibliography{reference}
%

%

\begin{IEEEbiography}[{\includegraphics[width=1in,height=1.25in,clip,keepaspectratio]{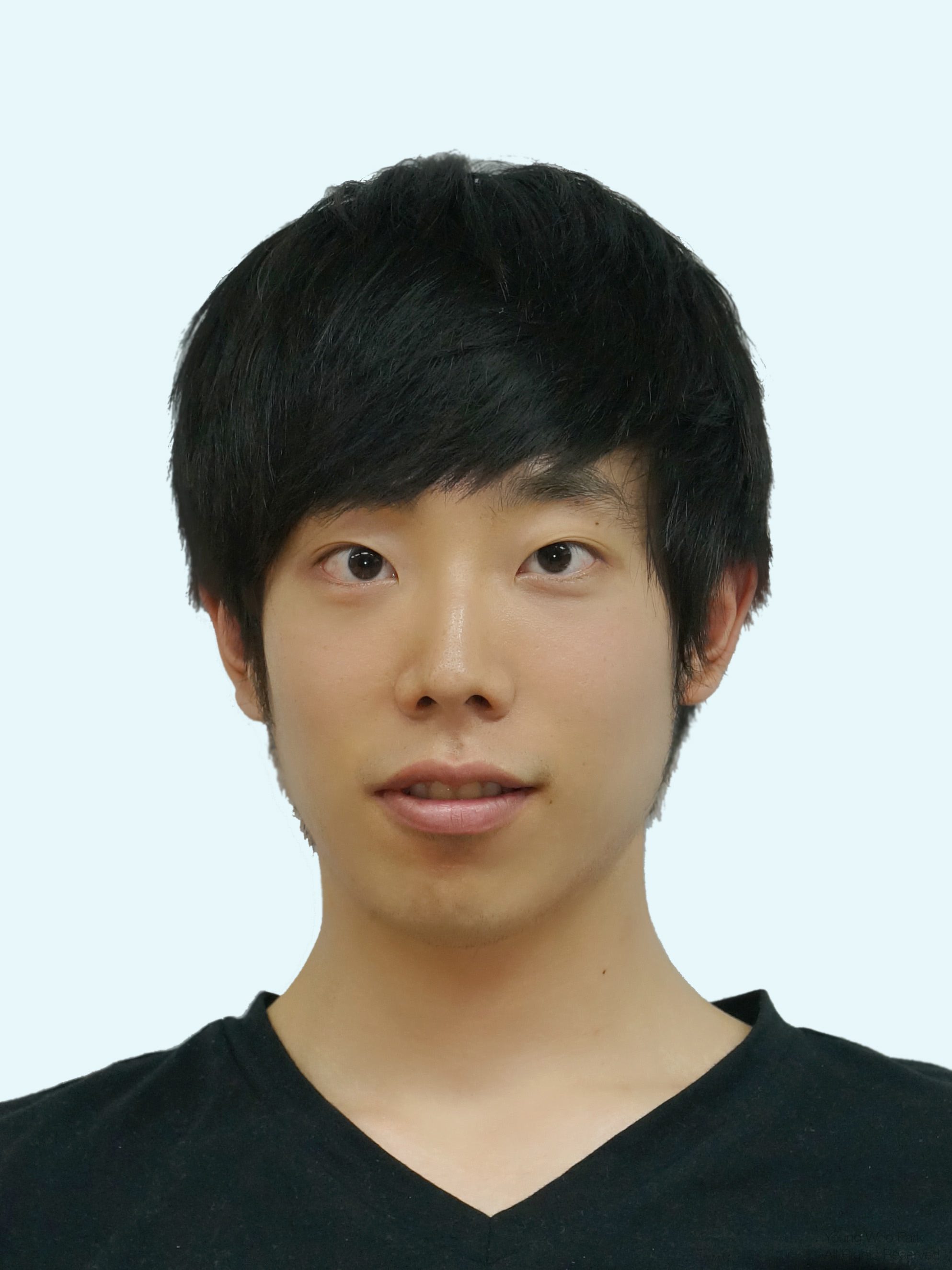}}]{Ue-Hwan Kim}
received the M.S. and B.S. degrees in Electrical Engineering from Korea Advanced Institute of Science and Technology (KAIST), Daejeon, Korea, in 2015 and 2013, respectively. He is currently pursuing the Ph.D. degree at KAIST. His current research interests include visual perception, service robot, cognitive IoT, computational memory systems, and learning algorithms.
\end{IEEEbiography}

\begin{IEEEbiography}[{\includegraphics[width=1in,height=1.25in,clip,keepaspectratio]{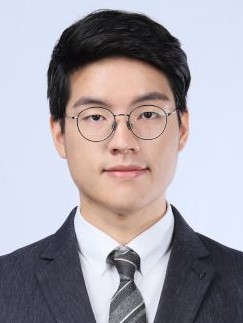}}]{Se-Ho Kim}
received the B.S. degrees in Electrical Engineering from KAIST, Daejeon, Korea, in 2019. He is currently pursuing the M.S. degree at KAIST. His current research interests include visual perception, computer vision, and learning algorithms.
\end{IEEEbiography}

\begin{IEEEbiography}[{\includegraphics[width=1in,height=1.25in,clip,keepaspectratio]{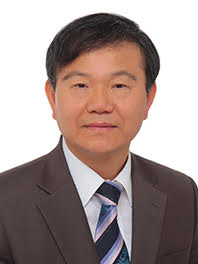}}]{Jong-Hwan Kim}
(F'09) received the Ph.D. degree in electronics engineering from Seoul National University, Korea, in 1987. Since 1988, he has been with the School of Electrical Engineering, KAIST, Korea, where he is leading the Robot Intelligence Technology Laboratory as KT Endowed Chair Professor. Dr. Kim is the Director for both of KoYoung-KAIST AI Joint Research Center and Machine Intelligence and Robotics Multi-Sponsored Research and Education Platform. His research interests include intelligence technology, machine intelligence learning, and AI robots. He has authored 5 books and 5 edited books, 2 journal special issues and around 400 refereed papers in technical journals and conference proceedings.
\end{IEEEbiography}





\end{document}